\documentclass{article}

\usepackage{PRIMEarxiv}

\usepackage[utf8]{inputenc} 
\usepackage[T1]{fontenc}    
\usepackage{hyperref}       
\usepackage{url}            
\usepackage{booktabs}       
\usepackage{amsfonts}       
\usepackage{nicefrac}       
\usepackage{microtype}      
\usepackage{lipsum}
\usepackage{fancyhdr}       
\usepackage{graphicx}       
\graphicspath{{fig/}}     

\usepackage{amsmath}      

\usepackage[sorting = none, autocite = superscript, backend = bibtex, style=numeric]{biblatex}
\let\cite=\supercite
\usepackage{listings}
\usepackage{xcolor} 
\usepackage{blindtext}
\usepackage{hyperref}
\usepackage{comment}
\usepackage{soul}

\newcommand*{\Win}{ ${\bf W}_\text{in}$  }
\newcommand{\Wres}{ ${\bf W}_\text{res}$ }
\newcommand{\Wout}{ ${\bf W}_\text{out}$ }
\newcommand{\rctorch}{\texttt{RcTorch}}
\newcommand{\rctorchS}{\texttt{RcTorch} }
\newcommand{\pytorch}{\texttt{PyTorch}}
\newcommand{\pytorchS}{\texttt{PyTorch} }

\newcommand{\botorchS}{\texttt{BoTorch} }
\newcommand{\hps}{\texttt{HPs}}
\newcommand{\rcbayes}{\texttt{RcBayes}}
\newcommand{\rcnetwork}{\texttt{RcNetwork}}
\newcommand{\turbo}{\texttt{TuRBO}}
\newcommand{\turboS}{\texttt{TuRBO} }

\newcommand{\uu}{ {\bf u}  }
\newcommand{\hh}{ {\bf h}  }
\newcommand{\s} { {\bf s}  }
\newcommand{\bb}{ {\bf b}  }
\newcommand{\cc}{ {\bf c}  }
\newcommand{\wout}{ {\bf W}_\text{out} }
\newcommand{\win}{ {\bf W}_\text{in} }
\newcommand{\wres}{ {\bf W}_\text{res} }

\begin{document}
\title{RcTorch  \\  \Large a pytorch reservoir computing package with automated hyper-parameter optimization}

\author{
  Hayden Joy, Marios Mattheakis, Pavlos Protopapas \\
  John A. Paulson School of Engineering and Applied Sciences \\ 
  Harvard University \\ 
  Cambridge, Massachusetts 02138, United States\\ 
  \texttt{hnjoy@mac.com, mariosmat@seas.harvard.edu, pavlos@seas.harvard.edu} \\
}

\maketitle

\begin{abstract}
 Reservoir computers (RCs) are among the fastest to train of all neural networks, especially when they are compared to other recurrent neural networks. RC has this advantage while still handling sequential data exceptionally well. However, RC adoption has lagged other neural network models because of the model's sensitivity to its hyper-parameters (HPs). A modern unified software package that automatically tunes these parameters is missing from the literature. Manually tuning these numbers is very difficult, and the cost of traditional grid search methods grows exponentially with the number of HPs considered, discouraging the use of the RC and limiting the complexity of the RC models which can be devised. We address these problems by introducing \rctorch, a \pytorchS based RC neural network package with automated HP tuning. Herein, we demonstrate the utility of \rctorch by using it to predict the complex dynamics of a driven pendulum being acted upon by varying forces. This work includes coding examples. Example Python Jupyter notebooks can be found on our GitHub repository \url{https://github.com/blindedjoy/RcTorch} and documentation can be found at \url{https://rctorch.readthedocs.io/}.
\end{abstract}

\keywords{Reservoir Computing \and Neural Networks \and Echo State Networks \and Bayesian Optimization \and PyTorch \and Control  Dynamical Systems \and Data Driven \and Machine Learning  \and AI}




\markboth{Hayden Joy, Marios Mattheakis, Pavlos Protopapas}
{RcTorch: a pytorch based reservoir computing package with automated hyper-parameter optimization}

\title{%
  \large RcTorch:
  a PyTorch Reservoir Computing package \\
   \normalsize with Bayesian hyper-parameter optimization}







\maketitle




\section{Introduction}


Reservoir computers (RCs), also known as echo state networks, are specialized, artificial neural networks. They are powerful and train very fast. However, today RC is not as commonly employed by the machine learning community at large as other classes of neural network models. Unlike most neural networks, the majority of RC weights are not optimized via the backpropagation algorithm. Instead, the weights are generated via a stochastic process that is very sensitive to a few numbers, typically less than 10, called hyper-parameters (HPs). In a simple feed forward neural network, rather than being optimized during training, HPs govern the learning process. The optimization of HPs is a significant challenge to the widespread adoption of the RC. Other classes of neural networks, which are extremely popular today, have faced similar significant challenges in the past. At times, pessimism about neural networks and, thus, AI more broadly, has led to aversion to these models by the scientific community at large, resulting in an AI winter: a “period following a massive wave of hype for AI characterized by a disillusionment that causes a freeze in funding and publications"\cite{kurenkov2020briefhistory}.


\vspace*{\fill}
----------------------------------- \\
Preprint. Under Review

\pagebreak

AI winters (both generally and for specific models) follow a general historical pattern. First, a new model is introduced that creates great excitement. Next, a technical barrier is introduced that causes the network to fall out of favor and stagnate. Finally, a solution is introduced, which removes the barrier. This leads to technical breakthroughs and, subsequently, to a flood of new papers, attention, and funding \cite{backprop}.

In this study, we attempt to remove barriers to use for RC by introducing \rctorch: a library for RC that makes reservoir computing easy and accessible. To the best of our knowledge, such a unified and maintained RC library with automated HP tuning does not exist. Our package is flexible and modern - it is written in the popular machine learning framework \pytorch. 

This manuscript has many sections which may be of varying interest to the reader. Section 1 gives an overview of the RC. Section 2 puts the RC in the context of the history of other neural networks, which is crucial to understanding their importance in modern machine learning, but a reader who is more focused on the specific technical details of the RC may skip this part and start reading from Section 3. The remainder of this work  demonstrates how to use \rctorch, as well as its powerful predictive ability, on a forced pendulum example.

\subsection{Introduction to the RC}
\label{section:rc_intro}

Reservoir computers are a specialized subset of recurrent neural networks (RNNs), so it is important to understand RNNs and their shortcomings. RNNs are a class of artificial neural networks where the hidden states at time-step $t$ depend on the previous hidden states. This establishes temporal dynamic behavior that yields a computation platform with an internal memory. The ability of RNNs to make predictions based on past events makes them powerful tools for handling temporal and sequential data. However, training an RNN is difficult and time-consuming due to its recurrent connections. Also, the problems of exploding and vanishing gradients are often observed.\cite{diffulty_rnns} RC was first introduced \cite{Jaeger2001} as a subset of RNN architectures that are considerably less computationally expensive to train. The RC consists of three layers: input, reservoir (hidden), and output (readout). Training the RC is efficient because only the weights connecting the reservoir to the output layer are updated during the network training (or fitting) phase. In the RC architecture, the hidden layer has randomly and sparsely connected neurons. These are collectively known as a `reservoir' of neurons.\cite{Jaeger2001} As opposed to the typical RNN, the connections of the neurons in the input layer and the recurrent hidden nodes in the reservoir layer of an RC are fixed. Because of these frozen weights, the RC can utilize a large number of hidden neurons. This in turn allows for sufficient complexity to develop an intelligent device.

 The simplicity of the RC architecture, and its substantially reduced learning cost compared to iterative learning algorithms, has attracted a large amount of attention in the machine learning and artificial intelligence (AI) communities. Even as early as 2012, RC had been employed in applications as diverse as robotic control, financial forecasting, and the detection of epileptic seizures \cite{reservoir_trends_2012}. More recently, it has been shown that unsupervised reservoir computers can solve ordinary differential equations \cite{mattheakis_unsupervised_2021} and that the use of ~``observer" time-series can assist in the analysis of chaotic data\cite{prl2018_ott, neofotistos_machine_2019, Ott} and to model turbulent flow.\cite{2Dturbulence}

One of the most interesting aspects of RC is that it can form physical hardware neuromorphic reservoir computers \cite{lugnan_photonic_2020}. This can lead to networks that are orders of magnitude faster than software-based computational methods. In recent years, several neuromorphic RC architectures have also been implemented in physical reservoirs on dedicated hardware, such as photonic integrated circuits \cite{scRep8_2018},  photonic cavities \cite{scRep9_2019}, optoelectronic technology \cite{prx7_2020},  and nonlinear wave dynamics.\cite{prl_2020} These architectures can potentially be deployed for real-time machine learning predictions. However, even in hardware RC  the discovery of high quality HPs remains a significant challenge.

\subsubsection{The crux of RC: Optimizing the hyper-parameters}
While RC networks train extremely quickly, there is no free lunch in optimization problems. That is, ``there is no optimization algorithm that can outperform all optimizers when solving all problems''.\cite{free_lunch1, free_lunch2, lightning_2016} The architecture of the RC is mostly determined by HPs that are not learned during the training process. The RC is extremely sensitive to its HPs. The difficult task of fine-tuning the optimal HPs, which often varies for different problems, has kept the RC a niche model mostly used by experts.  Conventionally, the HPs of RC were tuned by using a simple grid search. However, the computational cost of this method scales exponentially as the number of HPs increases. Given $n$ HPs, the space searched for the real-valued HPs is $n$-dimensional. Therefore, adding an HP involves adding a dimension to the search space. This is a significant challenge for a simple grid search, as the search time explodes exponentially as more and more HPs are added. In practice, grid search and manual tuning quickly becomes infeasible when $n > 3$, forcing the user to design relatively simple models. One solution to this problem is Bayesian optimization (BO).\cite{pavlos_BO} BO is a powerful method for optimizing the HPs of computationally expensive tasks such as neural network optimization. As Geoffrey Hinton suggested in a NeurIPS workshop 2016: ``[BO] replaces the graduate student who fiddles around with all these different settings for the hyper-parameters''.\cite{Hinton_lecture_2016}




\subsubsection{Bayesian Optimization and \rctorch}

In this study, we present a new RC library called \rctorch which creates RC networks and also has a powerful BO class capable of optimizing the RC HPs. \rctorchS is capable of training many networks in parallel and then evaluating their performance, that is, generating a score for each model. These scores are then reported to the RcBayes class which performs BO. Thus, \rctorchS generates many RC architectures and then selects the best one based on a robust and efficient process which can converge in as few as a hundred iterations. BO is an active field of research and is well documented in the literature.\cite{unk2019BO,  dai2020federated, Maat2018} In particular \rctorchS  utilizes a \pytorchS library called \botorchS which is concerned with BO techniques and was released by the Facebook AI Lab.\cite{Botorch2020}


\rctorchS employs the \turboS algorithm, which is a local BO method.\cite{TURBO2020, Botorch2020}  While global BO methods do not subset the HP search space, \turboS effectively divides different parts of the HP search space. Global methods tend to over-explore the search space, optimizing poorly in local regions. Moreover, \turboS can be parallelized and thus, it is faster and returns higher quality HPs than global BO methods. By employing \turboS BO, \rctorchS is able to optimize many HPs efficiently and can find better HPs than standard BO methods. More technical details about BO and \turboS in \rctorchS can be found in section \ref{section:BO}.

    Automated BO has the potential to make the RC a popular model by eliminating the tedious task of searching for optimal HPs. In the next section, to emphasize the significance of enabling widespread adoption of effective models, we give historical examples of the challenges of other neural networks. We also examine the outcomes of overcoming those challenges.


\section{A brief history: Cycles of AI winters}
\subsection{The first AI winter: Feed Forward Neural Networks (FFNNs) and their initial challenges (1958 - 1986)}	

The earliest neural network, the single-layer perceptron, was introduced by Frank Rosenblatt in 1958 and was greeted with hyperbolic fanfare.\footnote{The so-called perceptron, consisted of a single layer of simulated neurons.}\textsuperscript{,}\cite{kurenkov2020briefhistory} The New Yorker reported that \emph{“[Rosenblatt’s perceptron] is the first serious rival to a human brain ever devised”}. The New York Times claimed, in turn, that \emph{“The navy revealed the embryo of an electronic computer today that it expects will be able to walk, talk, see, write, reproduce itself, and be conscious of its own existence"}. The paper then reported that \emph{ ``[Rosenblatt] said Perceptrons might be fired to the planets as mechanical space explorers"}.\cite{kurenkov2020briefhistory} However, this overzealous excitement did not last.

While at first it appeared that Rosenblatt had created the first universal learning machine, by 1969 Marvin Minsky of MIT famously proved that Rosenblatt's perceptron failed to approximate the simple logical function ``exclusive or" (XOR). The perceptron more generally failed to solve functions that were not linearly separable. This was particularly devastating to the popular perception of AI because in the late 1950s and early 1960s, it was thought that solving AI essentially boils down to solving the rules of logic. Minsky specifically proved that a multi-layer model would be needed to solve this fundamental problem, but an algorithm capable of training such a network was not then known. The gap between dramatic proclamations of the promise of artificial intelligence and the limitations of the single-layer perceptron led to pessimism and aversion by the scientific community at large. Thus began the first AI winter.


\subsection{The spring of FFNNs and the rise of Recurrent Neural Networks (RNNs)}

In 1986 Rumelhart et al. \cite{backprop} popularized backpropagation (backprop) - the algorithm that solved the training of a multi-layer neural network. Backprop ended the first AI winter and brought attention back to neural networks. In that same article, the authors introduced the first RNN.\footnote{In his seminal 1975 thesis, Werbos first described the process of training neural networks through the backpropagation of errors. \cite{backprop}} RNNs have an additional set of neurons,  known as the hidden states, that pass information (a context vector) from one time-step to the next. This context vector can be viewed as the recurrent memory of the network. An RNN may also be seen as a Euler approximation of a time dependent ordinary differential equation (ODE) \cite{mattheakis_unsupervised_2021}, which gives a strong theoretical justification for their use in time dependent problems. However, RNNs can also be unwrapped and viewed as a very deep feed forward neural network, with as many layers as the time steps of the input data. More than being just an interesting technical perspective, this fact brings a host of challenges which were nearly debilitating during the RNN's inception in the 1980's and 1990's.

\subsubsection{The Vanishing Gradient Problem: The crux of RNNs}

Very deep networks are notoriously difficult to train due to the vanishing gradient problem\footnote{also known as the gradient vanishing/exploding problem}, which is the tendency of gradients to explode or shrink in deep neural networks. To perform gradient descent with backprop to find the optimal set of weights, we have to take derivatives. Backprop works by using the chain rule to calculate the gradient of the loss function with respect to the weights of the network. This is done from the final layer of the network back toward the input layer. For an RNN this means that the gradients must be calculated at every time step. Such a large application of the chain rule through a lot of layers amounts to a massive product - numbers slightly larger than one or less than one will tend to explode towards infinity or vanish towards zero, respectively.\footnote{Gradients vanish and explode when training RNNs because we must backprop back in time through their hidden states. This amounts to a very large application of the chain rule.  Given an error function $E$, a prediction vector $\mathbf{\hat{y}} = (y_1, ..., y_n)$ an output network weight $U$, and hidden states $h_1, h_2, h_k, ... h_n$ the derivative of the loss function with respect to the output weight is as follows:
$\frac{d L_t}{d U} = \sum_{k = 1}^t{\frac{\partial L_t}{\partial \hat{y}_t}\frac{\hat{y}_t}{\partial h_t}\frac{\partial h_t}{\partial h_k}\frac{\partial h_k}{\partial U}}$
where $\frac{\partial h_t}{\partial h_k} = \prod_{j = k +1}^{t}{\frac{\partial h_j}{\partial h_{j-1}}}$.
}  When a number is too large or too small, the computer cannot represent it with sufficient precision, and information is lost. If we had infinite precision in our instruments, this would not occur, but alas.  Thus, simple (or vanilla) RNNs struggle to train sequential data longer than 20 to 50 discrete time-steps.\cite{bengio_learning_1994}



\subsubsection{Three key properties of an effective time dependent model}
 In the same 1994 paper, where Bengio et al. \cite{bengio_learning_1994} performed one of the first rigorous analyses of the vanishing gradient problem in RNNs, they also laid out three necessary qualities of a good time-dependent model. \emph{\textbf{In this study we judge our AI software by these criteria}}. The model should: \emph{(1) be able to work over an arbitrary time duration, (2) be robust to noise, and (3) be capable of solving problems in reasonable amounts of time.} Bengio et al. showed empirically that vanilla RNNs failed all of these criteria even for simple problems. The RNNs were not ``efficiently trainable" with gradient descent, were not robust to noise, and could not solve problems for longer than 50 time steps.\cite{bengio_learning_1994} RNNs were initially thrown out in the cold due to these difficult challenges. Like the feed forward networks before them, RNNs were waiting for their day in the sun.

\subsubsection{Long Short Term Memory models and the Golden Age of RNNs (1997-2019)}

In 1997 the Long Short Term Memory (LSTM) (a class of RNN)  was proposed to address the vanishing gradient problem.\cite{lstm1997} The authors proved that by allowing the gradients to flow directly (via a linear activation and so-called ``memory gates'') RNNs could make longer predictions. While these models empowered RNNs to make longer predictions, the computationally expensive nature of their training (and the insufficient processing power of computers in the late 1990's and 2000's) caused them to fade. This represented the onset of the second AI winter, a period during the early 2000's when neural networks fell out of favor with the rise of decision tree models and support vector machines.


By the late 2000s, the second winter would begin to thaw and between 2010 and 2015 relentless interest and funding flooded back to neural networks. Hinton et al. employed GPU acceleration to empower deep neural networks (deep learning) and progressively beat records in computer vision and speech recognition. Soon his students were being hired at top tech companies.\footnote{By 2010 two of his students were hired at Microsoft and jump started their artificial intelligence research division. This represented big tech's first serious attempt to integrate neural networks in its products.} For example, in 2011 one of Hinton's pupils, Navdeep Jaitly, went to Google where he interned with the speech recognition team. Shortly thereafter Google's previous solution to speech recognition was thrown out in favor of a new neural network approach.\cite{kurenkov2020briefhistory} In 2015, Google used LSTMs to reduce speech transcription errors by 50\%. By 2016, they cut translation errors by 60\%.\cite{google_voice} Adoption of RNNs for natural language processing by other large tech companies soon followed. In 2017, Apple and Amazon announced separately that they had introduced LSTMs into Siri and Alexa, respectively, \cite{amazon, alexa_2018, apple}; Facebook and Microsoft soon followed suite \cite{microsoft, fb}. However, despite a meteoric rise in the language model space, these models still suffered from the gradient vanishing problem to a degree and slow training and since 2019 their use has declined, the natural language processing was dethroned by attention based models such as transformers.\cite{noauthor_deep_2021} While the RC does not suffer from the weaknesses of LSTMs, it has its own barriers to use.

\subsection{The current winter of the RC}

    For RC, the problem of the HPs has been a barrier similar to the barrier of training a multi-layer perceptron in the 1960s and 1970s or RNNs in the early 1990s. While there are many useful RC software packages, such as \texttt{pyESN} and others \cite{cknd_github, github_echotorch, github_echoes_2022, github_pytorch-esn_2022}, these packages lack automated hyper-parameter tuning capabilities.\footnote{An exception is the excellent library written by Reinier Maat, which was the first library to demonstrate that the HPs  of the RC could be efficiently tuned with BO. However, the packages upon which that library was built — gpyopt — is no longer maintained. \rctorch is written in PyTorch, considered by many to be the best AI language for researchers and is constantly maintained.\cite{why_pytorch_2020}} This makes using the RC feel more like guess-and-check work rather than machine learning. Typically, the researcher must use another software package or utilize an independent approach for the discovery of the optimal \hps.
     \space \rctorchS seeks to remove the barrier of the difficult task of hyper-parameter optimization with automated HPs tuning via BO.  Due to the crucial role the HPs play in determining the architecture of the RC network, it is appropriate to introduce them in that context.
    
\st{\subsection{Background: RC as a subset of RNNs}}
    
\section{The RC Architecture}
\subsection{The general architecture of the RC}

The critical HPs  of the RC are related to its architecture. We adopt a similar notation as that used in by Ott. et al.\cite{Ott} The RC is made up of three sets of weights: input weights \Win with $M$ input nodes, reservoir weights \Wres with $N$ hidden (reservoir) nodes, and output weights \Wout with $P$ output nodes. In particular, the total set of weights $\mathbf{W}=\{ \win, \wres, \wout\}$ where $\win \in \mathbb{R}^{N \times M}$, $\wres \in \mathbb{R}^{N \times N}$, and $\wout \in \mathbb{R}^{N \times P}$. While the application in this study is time dependent, the input and output to the RC can more generally be any sequential data.

RC is a subset of RNNs; thus, the following formulation describes any RNN of this form. We consider sequential data that comprise $M$ different sequences with $K$ sequential data-points, hence we read  an input vector $\uu_k \in {\rm I\!R}^M$ where $k$ is the sequential index.  The input vector feeds an RNN that consists of $N$ hidden recurrent states $\hh_k
\in {\rm I\!R}^N$ that form a reservoir of hidden neurons. 
The evolution of the hidden states in this study is given by:
\begin{align}
\label{eq:ht}
\hh_k = \left(1-\alpha \right)\hh_{k-1} + \alpha f\left( \wres \cdot \hh_{k-1} +\win\cdot \uu_k +\bb  \right),
\end{align}
where the dot indicates the inner product operator,  $\alpha$ is a {\it  leaking rate} that controls the evolution of the hidden states (smaller the $\alpha$ slower the evolution),  $f(\cdot)$ is an arbitrary nonlinear activation function, the matrix $\wres\in {\rm I\!R}^{N\times N}$  is an adjacency matrix that contains the weights of the connections between the hidden neurons, and $\win\in {\rm I\!R}^{N \times M}$,  $\bb\in {\rm I\!R}^{N\times 1}$ denote, respectively, the weights and biases of the input layer. \footnote{This is but one implementation of an RNN. Hidden states could more generally be represented by $\hh_k = \phi \left( \wres,  \hh_{k-1}, \uu_k\right)$ where the function $\phi$ represents a more general function of hidden states and inputs. RC or RNN formulations can be more complex than the simple Gated Recurrent Network like formulation of the RC imposed herein.}


The network is performing a regression task and predicts a vector of $P$ sequences $\hat \s_k \in {\rm I\!R}^P$ that depends linearly on the recurrent states  through the relationship described by  Eq.~(\ref{eq:output}) where $\cc$ denotes a constant bias vector.\footnote{
More generally given an invertible output activation function $f_{out}$ the output layer can be expressed by  Eq.~(\ref{eq:output_generalized}). $\wout \in {\rm I\!R}^{N\times P}$ and  $\cc \in {\rm I\!R}^{1\times P}$   denote, respectively, the weights and biases of the output readout layer of the network.
\begin{align}
\label{eq:output_generalized}
f_{\text{out}}^{-1} \left(\hat \s_k\right) =   \wout \cdot \hh_k  + \cc,
\end{align}
}

\begin{align}
\label{eq:output}
\hat \s_k =   \wout \cdot \hh_k  + \cc.
\end{align}

\noindent Equations (\ref{eq:ht}),  (\ref{eq:output}) define forward propagation through an RNN where, in principle, all the weights and biases are trainable parameters. 

\begin{figure}[ht]
    \centering
    \hspace*{-.8cm}
    \includegraphics[scale=0.18]{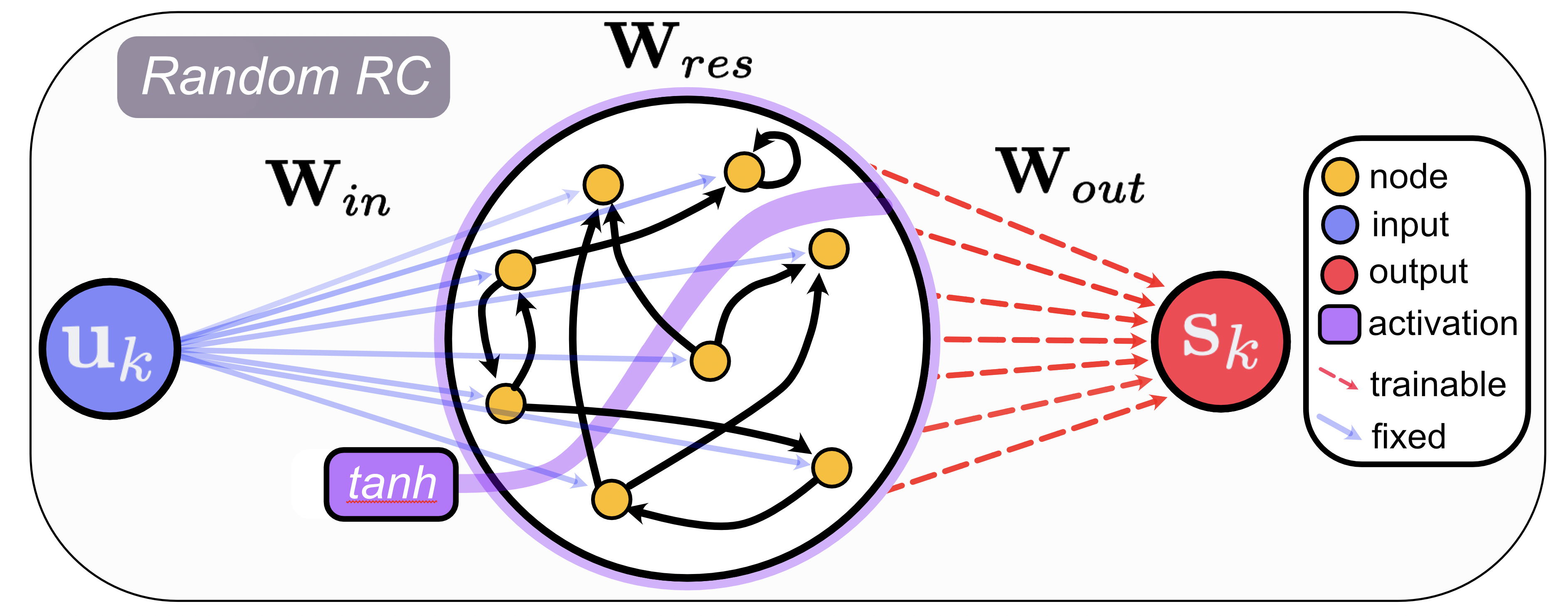}
    \caption{The standard RC architecture. The input $\uu_k$ is projected to a high dimensional vector space via \Win described by the hidden vector $\mathbf{h}_k$ (not shown) that evolves in time $t_k$.}
    \label{fig:second}
\end{figure}


The transition from a standard RNN to an RC occurs when we fix the parameters of the input and hidden layers while we train only the weights and biases of the output layer. The loss function that needs to be minimized in the training phase is given by:
\begin{align}
    \label{eq:loss}
    L = \sum_{k=1}^K\left( \hat \s_k -\s_k \right)^2 + \beta\text{Tr}\left[\wout \wout^T \right],
\end{align}
where $\s_k\in {\rm I\!R}^P$ is a vector that contains the ground truth data used to train the network.  The second term in Eq. (\ref{eq:loss}) is the ridge regularization penalty  \cite{Ott} used to avoid overfitting, and  $\beta \geq 0$ is the ridge regularization coefficient.\footnote{The regularization parameter can also be solved for via cross validation during training. This capability is under development for \rctorch.}

%
Having to optimize only the output linear layer, the network optimization simplifies to a ridge regularized linear regression problem. Considering that this problem has a closed form solution \cite{Jaeger2001, Ott, tino2020}, RC networks can be trained extremely quickly, making RC a very efficient RNN.

Let the input and output at time-step $k$ be $\mathbf{u_k}$ and $\mathbf{s_k}$. Fitting an \rctorch model consists of two steps: 1) constructing the hidden states and 2) training the output weights. Like a support vector machine, the RC uses its input weights to make a high dimensional projection of the input vector. The hidden state vectors $\mathbf{h_1}, \mathbf{h_2}, ..., \mathbf{h_k}$ can be considered as a non-linear and non-orthogonal set of basis functions which the RC linearly combines to approximate the target (response) variable. The HPs, with the exception of the ridge regression coefficient, determine the architecture of the hidden states, which are not learned by the network. During BO, we progressively search for different HP sets $S_i \in {\rm I\!R}^{|S_i|}$. Crucially, the weights of the reservoir are randomly chosen from distributions that are determined by some of the HPs. 

\subsection{HP overview}

An overview of the main HPs used in this study is given by table ~(\ref{tab:hyper}). $N$ represents the total number of nodes in the reservoir. The spectral radius $\rho$ is the maximum eigenvalue of the adjacency matrix (the adjacency matrix determines the structure of the reservoir). The hyper-parameter $\zeta$ is the connectivity of the adjacency matrix. The bias $b_0$ used in the calculation of the hidden states and the leakage rate $\alpha$ controls the memory of the network, i.e. how much the hidden state $h_k$ depends on the hidden state $h_{k-1}$. The ridge regression coefficient $\beta$ determines the strength of regularization at inference when solving, in one shot, for $\wout$.





 \begin{table}[ht]
 \centering
    \label{tab:hyper}
    \begin{tabular}{l|l|l}
    \textbf{HP} & \textbf{Description} & \textbf{Search Space} \\
    \hline
       $N$ & number of reservoir neurons &  typically 100 to 500\\
       $\rho$& spectral radius max eigenvalue of \Wres & [0,1]\\
       $\zeta$ & connectivity of the reservoir (1 - sparsity) & logarithmic \\
       $\mathbf{b_0}$&   bias used in the calculation of $\mathbf{h_k}$ & [-1,1]\\
       $\alpha$ &  leakage rate & [0,1]  \\
       $\beta$& ridge regression coefficient & logarithmic\\ 
    \end{tabular}
\end{table}

Although table ~(\ref{tab:hyper}) summarizes the HPs used in this study, \rctorchS is adaptive framework and more HPs can be easily hard-coded in \rctorch code. For example, an \rctorchS user might want to add non-linear dense layers on top of the RC. If the user wants to add multiple non-linear dense layers to the RC then a one-shot solution for the best output weights is no longer possible and gradient decent must be performed. In this case, the learning rate $\lambda$ would be an additional HP that could be optimized with BO.




 
 

\section{Software application of \rctorch: the forced pendulum}
We demonstrate the effectiveness of the proposed \rctorchS AI tool through an application; we solve the forced pendulum problem. An undamped driven pendulum is represented by the system of non-linear ordinary differential equations represented by Eq.~(\ref{eq_fp_x}) and Eq.(\ref{eq_fp_rho}) where $x$ is the position of the mass of the pendulum assumed to be one and $p$ represents its momentum. 

\begin{equation}
\dot{x} = p
\label{eq_fp_x}
\end{equation}

\begin{equation}
\dot{p} = - \sin(x) + \alpha f(\omega, t)
\label{eq_fp_rho}
\end{equation}

\begin{figure}[ht!]
    \centering
    \hspace*{-.8cm}
    \includegraphics[scale=0.45]{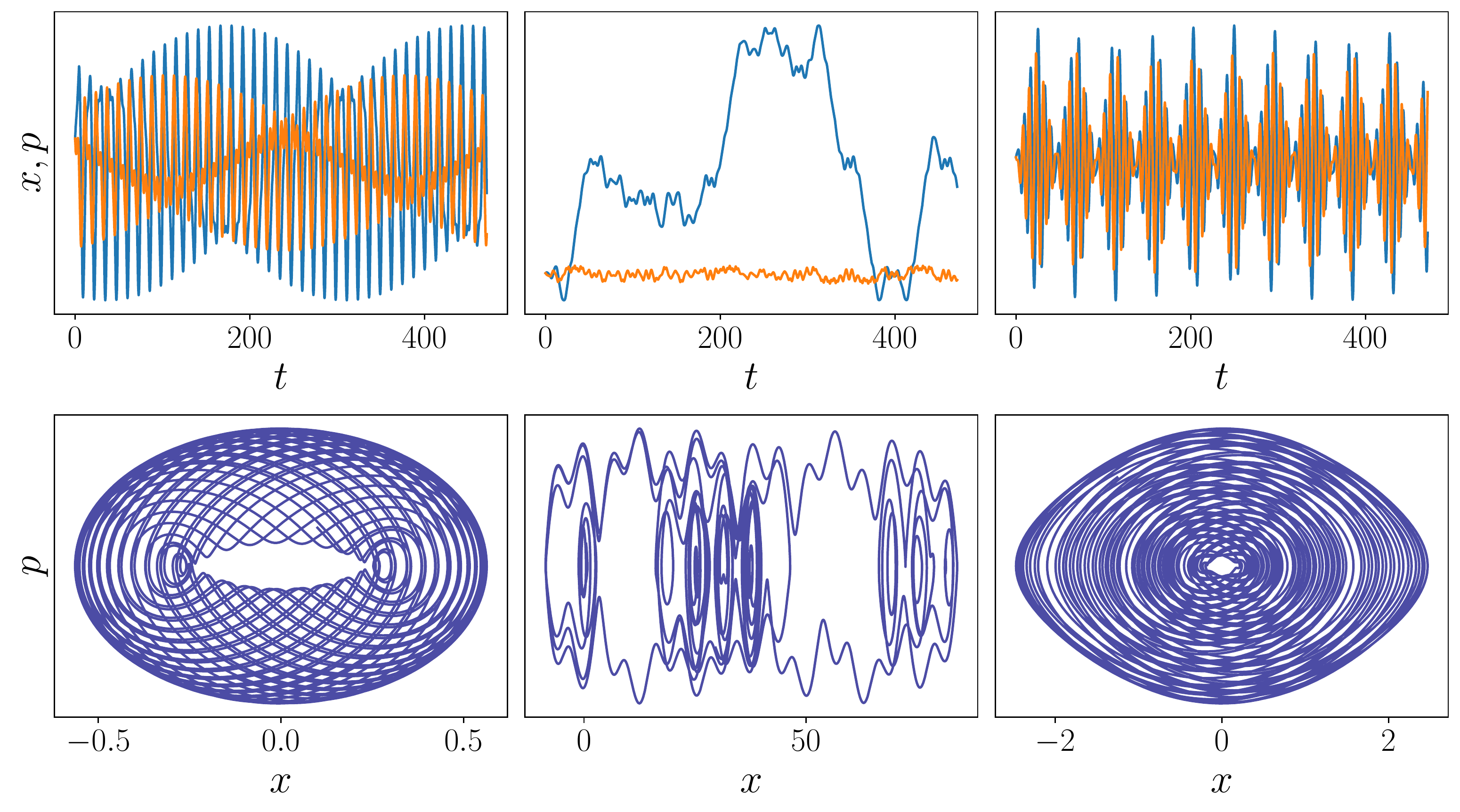}
    \caption{Top row: Example trajectories of the forced pendulum generated with the \texttt{odeint} package which employs LSODA algorithm from the FORTRAN library odepack.\cite{petzold_automatic_1983} Here, the force is defined by $\alpha \sin(\omega x)$. Plotted in blue and orange are the position $x(t)$ and the momentum $p(t)$, respectively. Bottom row: The momentum is plotted versus the position called the phase-space. The middle trajectory is resonant where the motion is not clearly bounded. From left to right the values of ($\alpha$, $\omega$) are [(0.3, 0.5), (0.5, 0.85), (0.2, 0.95)]. These plots demonstrate that the forced pendulum can behave in wildly different ways when only varying the force applied. For all sets the value of the initial conditions are $x_0 = 0.5$, $p_0=0.5$. }
    \label{fig:example_trajs}
    
\end{figure}

In Eq. (\ref{eq_fp_rho}), $ f(\omega, t)$ is a real value external driven force function bounded within [-1, 1], where $\omega$ and $\alpha$ are, respectively, the frequency and amplitude of the force. The forced pendulum can display wildly different behavior depending on the external driven force.  Fig.~\ref{fig:example_trajs} shows trajectory plots (above) and phase space plots (below) for various $\alpha f (\omega, t)$. Vertical pairs of plots correspond to a particular force.




%

For small initial position and velocity, the pendulum is a nonlinear periodic system, but for certain sets of $\omega$ and $\alpha$ the system begins to behave in a more complex way. The upper and lower middle panels of the second of Fig.~\ref{fig:second} demonstrate this behavior. Note that all data used in this study were generated using the \verb|scipy.integrate.odeint| implementation of the Fortran package written in Python.\cite{odeint}



The RC can perform two types of tasks: ``pure prediction" and ``parameter aware prediction". In the former case, the model does not have any information about the future dynamics. A standard example is the prediction of a stock price. This is in contrast to the parameter-aware case where some physical information or future variables of the system can be inferred. For example, in many robotics applications, the user knows the driven force applied by the robot, which is a given parameter that helps predict how the system will react to the force.

In our software demonstration, in the case of a driven pendulum, we first perform pure prediction with the RC using \rctorch. Pure prediction is when the RC takes time as input and is trained on the ground truth ($x$, $p$). Then, we improve upon this result with a parameter-aware RC. That is, the driven force is assumed to be a known function. Next, we show that \rctorchS can recover the underlying dynamics of the system even from noisy data. Finally, we demonstrate a transfer learning application of the RC: After learning one set of HPs optimized for a particular input force, our system can learn with a high degree of accuracy how the system will respond to other forces.

This software application builds up to demonstrating that for non-resonant frequencies, \rctorchS can predict the forced pendulum trajectories for {\em forces outside the training regime}. This means that the RC is capable of learning the general underlying dynamics of physical systems. Thus, the HPs of the RC can be viewed as encoding a rich latent embedding: a high-dimensional general representation of a system.

\begin{figure}[ht!]
     \hspace*{-0.77cm}
    \makebox[\textwidth][c]{\includegraphics[width = 0.9\textwidth]{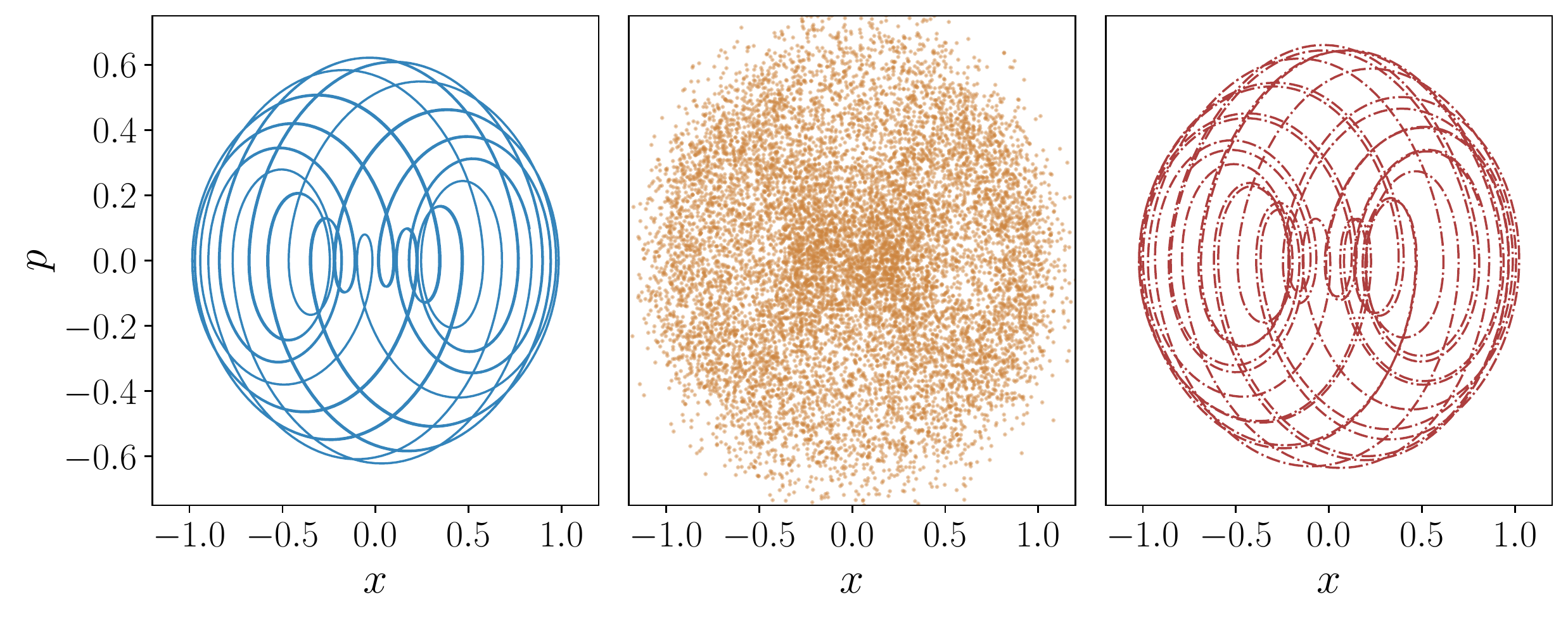}}
    \caption{Left (blue): The ground truth phase space of the forced pendulum. Center (gold, dotted): The phase space of the data, which was noisy and hard to discern. Right (red, dashed-line): the phase space recovered by the RC trained only on the noisy data.}
    \label{fig:hype}
    
\end{figure}

 Figure~\ref{fig:hype} shows three cases of a particular trajectory of the forced pendulum: the ground truth relationship, noisy data, and the RC prediction of a particular trajectory. All three plots are shown in phase space; where the position along the position ($x$) plotted against the momentum ($p$). Figure~\ref{fig:hype} demonstrates that the RC recovered the dynamics of the system. However, when the force is different enough from the original force, the model can begin to struggle, and new HPs should be found. We demonstrate how to find new HPs in section \ref{section:BO} which shows how to use the \rcbayes \space class to perform BO. Following the same procedure, a practitioner can recover the underlying dynamics and forecast dynamic behavior given noisy sequential data.

\section{Software Application Part 1: Forecasting}

In this section we demonstrate how we can use \rctorchS to solve the forced pendulum. Note that $>>>$ represents a command being entered by the user.

\subsection{ \rctorchS resources and installation}

The \rctorchS github repo can be visited at \url{Github repo: https://github.com/blindedjoy/RcTorch}.
In addition \rctorchS has clean and effective documentation that can be accessed at the link at \url{ https://rctorch.readthedocs.io}.

\rctorchS can be easily installed via pip:

\begin{center}
\begin{tabular}{c}
\begin{lstlisting}

>>> pip install rctorch

\end{lstlisting}
\end{tabular}
\end{center}


\subsection{Pure prediction with \rctorch}


A pure forecasting problem is a task where when making a prediction we do not have any knowledge of concurrent information (technically no time-series concurrent to the target). An example of pure prediction, as previously mentioned, is the prediction of a single stock price prediction as concurrent information is in the future and is inaccessible.

\subsubsection{Setting up the HPs}
\label{section:hps}


To get started with the software we use the following HP set which acts as a recipe for the RC.

\begin{center}
\begin{tabular}{c}
\begin{lstlisting}

>>> import rctorch
>>> hps = {`connectivity': 0.4071449746896983,
 `spectral_radius': 1.1329107284545898,
 `n_nodes': 202,
 `regularization': 1.6862021450927922,
 `leaking_rate': 0.009808523580431938,
 `bias': 0.48509588837623596}
 
\end{lstlisting}
\end{tabular}
\end{center}

This HP set was optimized via BO for $f=\alpha*\sin(\omega x)$ with $\alpha = 0.5$ and  $\omega = 0.2$. In section \ref{section:BO} we show how to search for an optimal HP set.

\subsubsection{\rctorch: Declaring an RcNetwork object}

Given this HP set we seek to generate a phase plot like Fig.~(\ref{fig:second}). In order to do so we demonstrate how to build, train, and test an  \rctorchS \rcnetwork \space class model. Classes in python are declared with function notation, so in order to build a specific instance object of the \rcnetwork \space class, we run the following line of code:

\begin{center}
\begin{tabular}{c}
\begin{lstlisting}
>>> my_rc = RcNetwork(**hps, 
                      random_state = 210, 
                      feedback = True)
\end{lstlisting}
\end{tabular}
\end{center}

Upon initialization, the \rcnetwork \space class object builds the reservoir and other necessary matrices. The HPs are passed as an unzipped dictionary to the \rcnetwork \space class. The argument \texttt{random\_state} ensures that the random stochastic (statistical) processes generating the network architecture are the same from realization to realization. This allows us to search for stable and optimal architectures. If this was not done, we would generate a different network architecture every single time we ran the code even when the HPs stayed consistent. The feedback argument is crucial for longer predictions as it passes the output of the network at time-step $t$ as an input to the network at time $t+1$, which allows the network to make longer and better predictions in general; this is also known as teacher-forcing. Next, we fit our model on unseen data.

\subsubsection{\rctorch: Fitting an RC}

When the fit method is called, \rctorchS iterates over time while building the hidden states using Eq. (\ref{eq:ht}). Once this is completed, the network solves for the optimal output weights for the last layer of the network.

\begin{center}
\begin{tabular}{c}
\begin{lstlisting}
>>> my_rc.fit(y = target_train)
\end{lstlisting}
\end{tabular}
\end{center}

In \rctorchS $y$ represents the target time-series which the network is supposed to predict and $X$ represents the input time series. In this case we did not need to enter $X$ because as we are doing a pure prediction task (the default value of $X$ is $None$, meaning that the only input is time).\footnote{Technically by default if $X$ is not provided to \rctorch, it will default to a vector of ones ($X = [1, 1, 1, ...]$), though using discrete time-steps $X=[0.1, 0.2, 0.3, ...]$ is also effective.} Next we test our model on unseen data. Here $y$ is described by the system of equations $\dot{x} = p$,  $\dot{\rho} = -sin(x) + 0.5 \sin(0.2 x)$ with no noise added. This is achieved by concatenating the two vectors.

\subsubsection{\rctorch: Testing the model}

\begin{center}
\begin{tabular}{c}
\begin{lstlisting}
>>> score, prediction = my_rc.test(y = target_test)
\end{lstlisting}
\end{tabular}
\end{center}

The  \texttt{test} method fist runs the \texttt{predict} method and subsequently calculates the mean square error (MSE) of the prediction versus the ground truth. The user can pass any other custom loss function by modifying the \texttt{criterion} argument of the test method. \footnote{Although the \texttt{test} method takes \texttt{target\_test} as an argument to score the result, \texttt{target\_test} is not passed to the \texttt{predict} method}. The \texttt{test} method returns a tuple of the overall loss (the score) and the prediction. Next, we can run the \texttt{combined\_plot} method to see our model predictions.

\subsubsection{\rctorch: Plotting RC predictions and residuals} 

\begin{center}
\begin{tabular}{c}
\begin{lstlisting}
>>> my_rc.combined_plot()
\end{lstlisting}
\end{tabular}
\end{center}

\begin{figure}[ht!]
  
    \noindent
    \makebox[\textwidth][c]{\includegraphics[width = 1\textwidth]
    {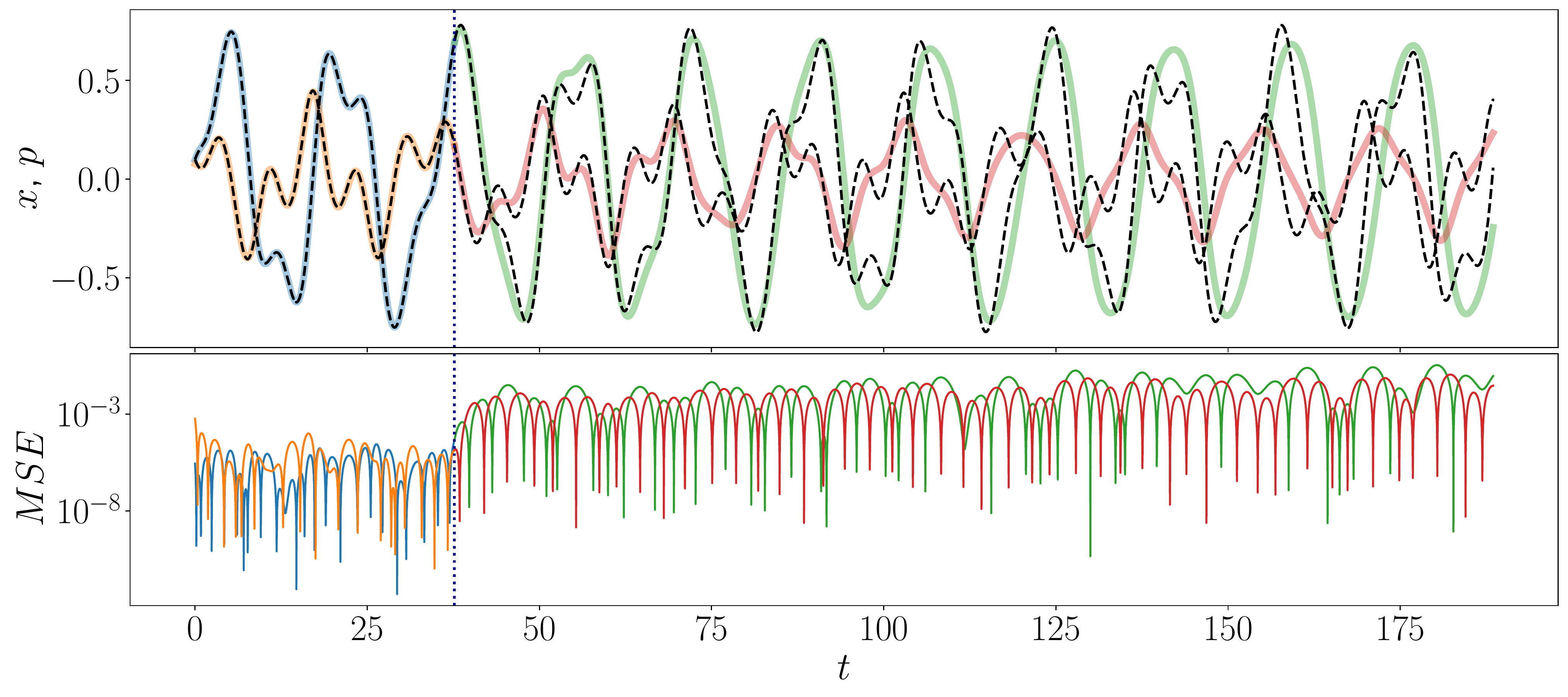}
    }
    \caption{The combined\_plot method plots both the prediction (above) and the log-residuals (below) versus time. The dotted blue vertical line separates the training and testing regimes. In yellow and blue on the left hand side of the prediction plot we see the position and momentum of the forced pendulum training sets and in green and red we see the RC predictions. The ground truth data for $x$ and $p$ are plotted as dashed black lines.}
    \label{fig:pure_prediction1}
\end{figure}


 While Fig. \ref{fig:pure_prediction1} demonstrates that the model did well in capturing the general timing of large wave-like fluctuations, \rctorchS is capable of better performance. This is despite the fact that the result is impressive within the context of the 20\%-80\% train-test-split and the fact that the data were a very long sequence of 12000 discrete time steps. The RC works with data over a long duration (many discrete time steps) and trains quickly, which is not true for other RNNs, which struggle over long sequences and are generally slow. The overall test set MSE\footnote{
$MSE=\frac{\sum_{k}{\left(s_{k}-\hat{s}_{k}\right)^2}}{\sum_{k}(s_{k})^2}$
} ($MSE$)
$\approx 0.141$ and the model took 7.93 seconds not including plotting time. All experiments were run on a MacBook Pro (2019) with a 2.4 GHz 8-Core Intel Core i9 processor and 64 GB 2667 MHz DDR4 of memory. All code was run in Python Jupyter notebooks without access to GPUs. 
 


\begin{figure}[ht!]
  \centering
  \hspace*{-0.8cm}
\includegraphics[scale=0.5]{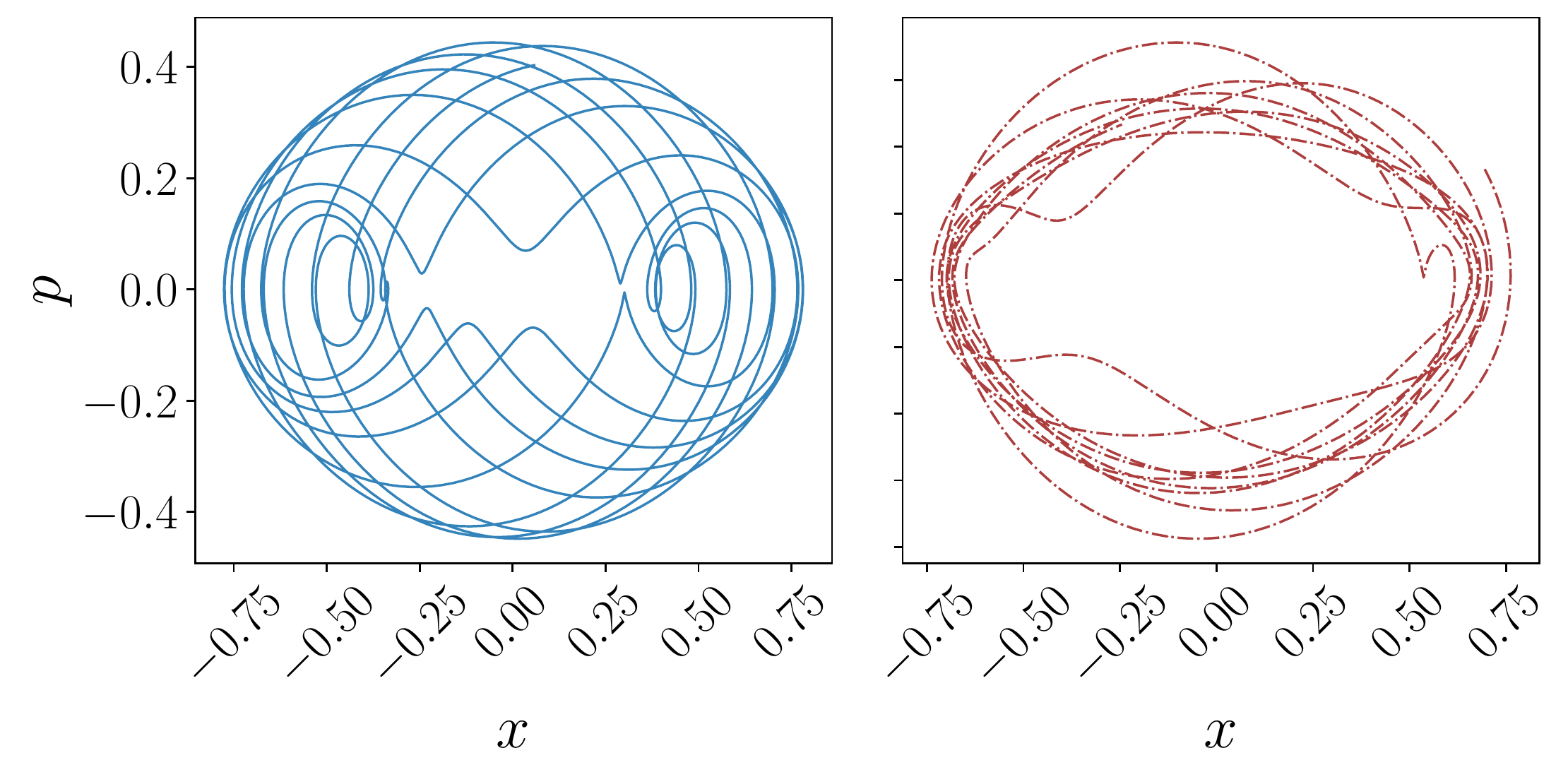}
\caption{Phase space plots; $x$ (position)  is plotted against $p$ (momentum). The real data is plotted in the left panel with a solid blue line. The \rctorchS \rcnetwork \space prediction is plotted at right with a dashed red line.}
\label{fig:pp_phase}
\end{figure} 


\subsection{Making a Parameter aware RC with \rctorch}

To improve the model, the force applied to the pendulum can be fed to the RC as input. In this ``parameter-aware" case, the force is assumed to be known. In control theory and robotics such known forces are called ``observers".\cite{Ott} \rctorchS can learn the mapping from the force  to the dynamics, namely to $x$ and $p$ of the system. Note that $X$ should be two dimensional (either a \texttt{numpy.nd.array} or a \texttt{PyTorch} tensor).\footnote{ This is the reason for the capital $X$. By convention, uppercase letters represent matrices, and lowercase letters represent vectors.}

\begin{center}
\begin{tabular}{c}
\begin{lstlisting}
>>> my_rc = RcNetwork(**hps, random_state = 210,  
            activation_function = "tanh", feedback = True)
>>> my_rc.fit( X = input_train, y = target_train)
>>> score, prediction = my_RC.test(X = input_test,  y = target_test)
\end{lstlisting}
\end{tabular}
\end{center}

\begin{figure}[ht!]
  \centering
  \hspace*{-0.8cm}
\includegraphics[scale=0.5]{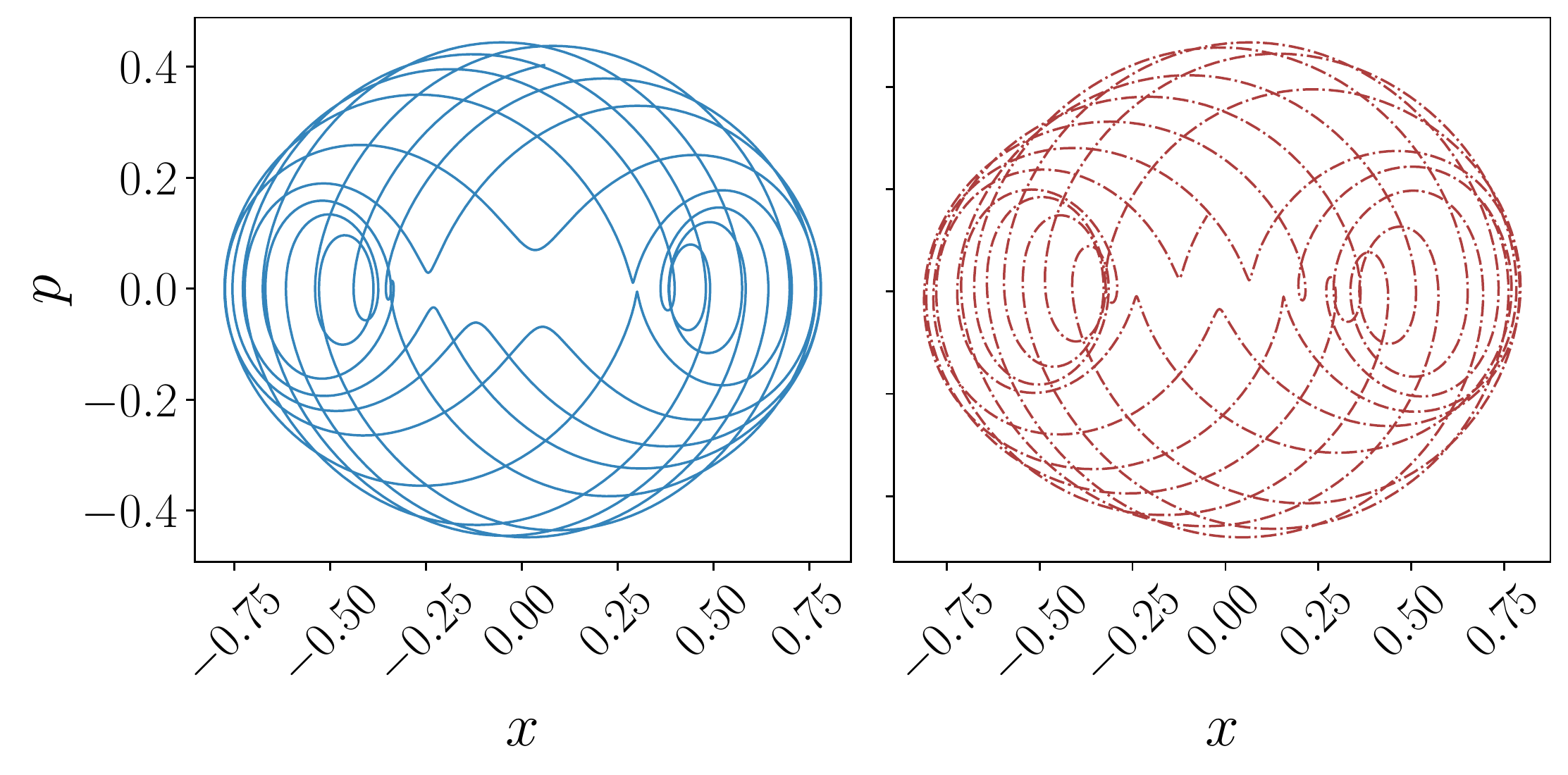}
\caption{ In this set of phase plots the parameter aware RC prediction (right) more closely resembles the ground truth (left) than in the pure prediction case. }
\label{fig:obs_phase1}
\end{figure}

 As demonstrated by Fig.~\ref{fig:noise_phase} the parameter aware RC  captured the dynamics of the system much better than the pure-prediction RC. This is demonstrated, in particular, by the more accurate forecast of the underlying phase-space. The overall MSE reported on the test set was $\approx 0.085$ and the model took 9.53 seconds to fit the data.

\subsection{Modifying activation functions with \rctorch}
\subsubsection{Output activation functions}
Choosing an appropriate activation function is an efficient way to improve the performance of a neural network 
(such as an RC). For example, we might make the assumption that the forced pendulum problem under consideration is a bounded problem. There are a host of other problems where bounding the output is reasonable, from planetary motion to the movement of a robotic arm. Other systems must obey the conservation of energy.\cite{mariosHNN} \rctorch \space can apply an activation function to the final layer of the network, which is by default the identity function. Because the range of the prediction seems to be bounded above and below, we can apply the hyperbolic tangent ($\tanh$) as its output is bounded between -1 and 1.\footnote{$\tanh$ is an invertible function. Therefore, we can use it as an output activation function, provided that the data that are fed to the RC lie in the domain [-1,1]. \rctorchS automatically scales the data for the user so that this is true, provided the training set and the testing set lie in similar ranges.} 

\subsubsection{Multiple activation functions}
\rctorchS allows the user to control the activation functions used throughout the RC, not just in the output. In the following code snippet, multiple activation functions were used instead of one. \rctorchS employs multiple activation functions to make hidden states more expressive. Namely, we get more complex dynamics of the states with more activation functions. \footnote{We ran a rigorous study of whether multiple activation functions or the output activation function is more critical to finding a lower loss and found that overall having varying activation functions was more impactful. This work can be found in the Appendix. In fact we used a special version of $ReLu$ which is only linear when for some input $x$, $|x|\leq\theta^{*}$  and otherwise is $\tanh$.}


\begin{center}
\begin{table}[ht]
\begin{tabular}{c}
\begin{lstlisting}
>>> my_rc = RcNetwork(**hps, random_state = 210,  
            feedback = True,
            output_activation = `tanh',
            activation_function = {`tanh' : 0.1, 
                                   `relu' : 0.9, 
                                   `sin': 0.05})
\end{lstlisting}
\end{tabular}
\end{table}
\end{center}

Above, \rctorchS assigns each reservoir node with an activation function with a probability weight proportional to the value of the dictionary above ($90\%$ for $ReLu$, $10\%$ for $tanh$, $5\%$ for $sin$). \rctorchS re-normalizes these probability weights internally so that they sum to one.

\begin{figure}[ht!]
  \hspace*{-0.7cm}
  \noindent
    \makebox[\textwidth][c]{\includegraphics[width = 1\textwidth]
  {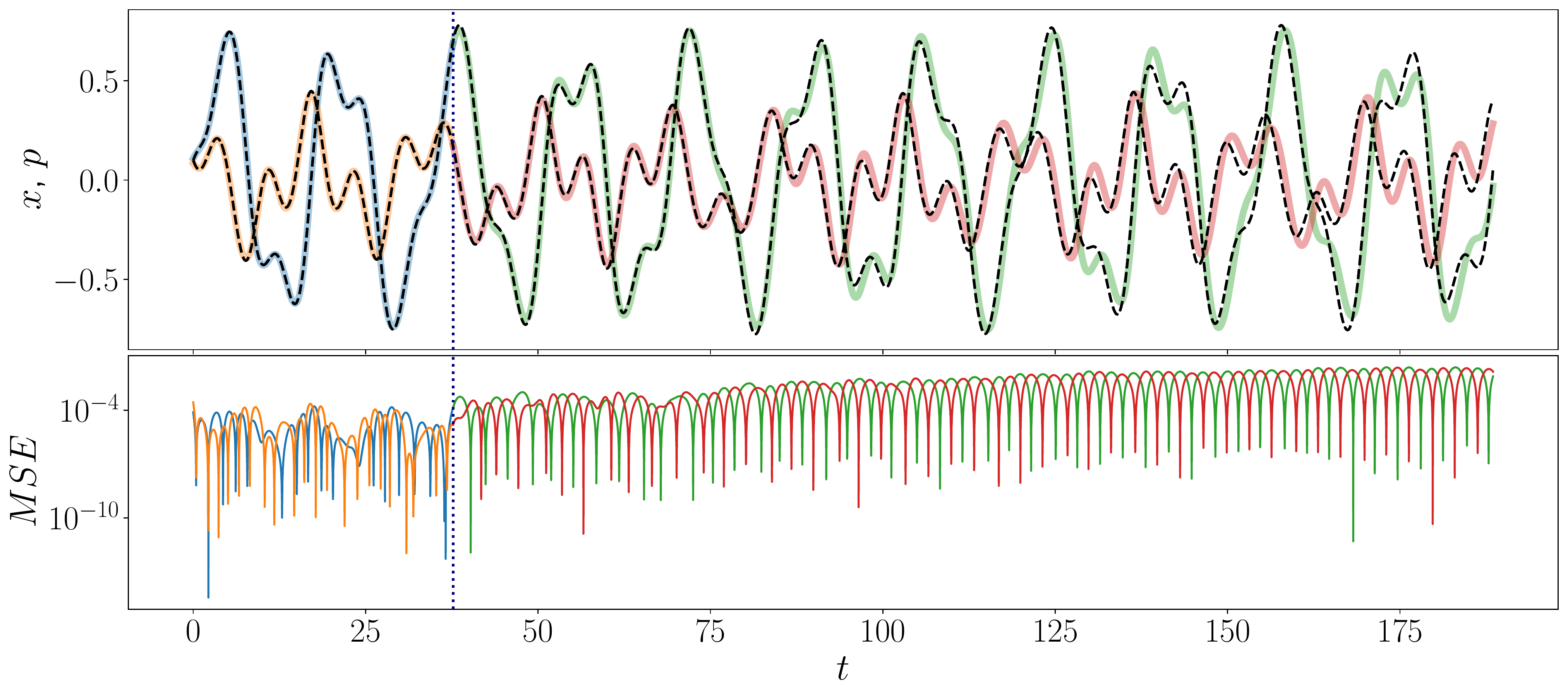}
  }
\begin{center}
\includegraphics[scale=0.5]{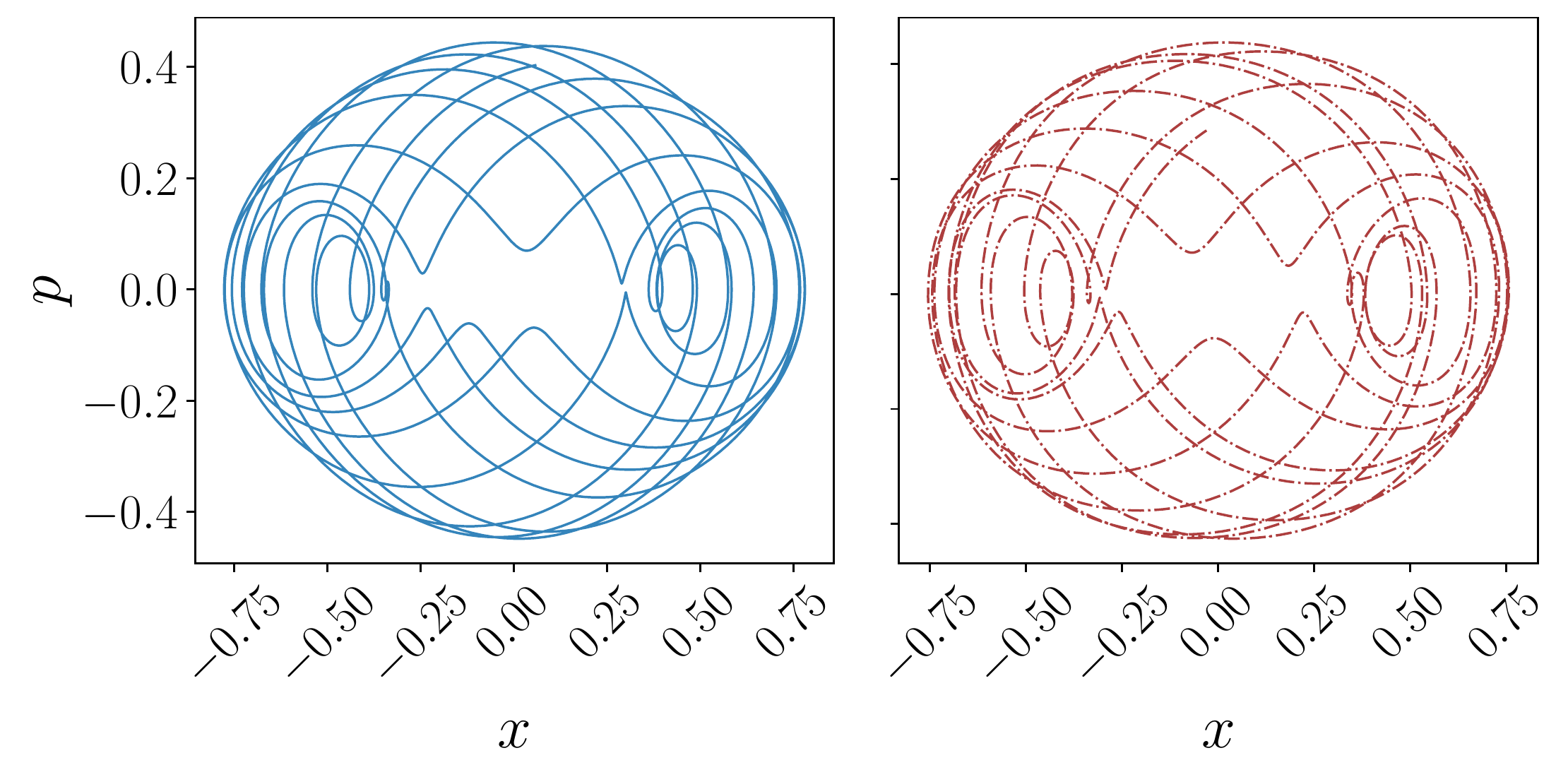}
\end{center}

\caption{From top to bottom: trajectory plot, middle residual plot, and phase-space plots plots. The ground truth phase space plot is at the bottom left and the RC prediction is at bottom right. After using multiple activation functions the non-linear non-orthogonal set of time dependent basis functions that make up the hidden states became more expressive and thus the prediction of the RC is even closer to the ground truth than the combined plot method output shown in Fig \ref{fig:obs_phase2}. However, because both predictions are excellent, this difference is subtle.}
\label{fig:obs_phase2}
\end{figure}

The trajectory and phase-space plots for this experiment are shown in Fig.~\ref{fig:obs_phase2}. The fit is very accurate. The experiment took 11.4 seconds to train with MSE $= 0.005$, or about 30 times less than the error in the first RC fit, the result of which is shown in Fig.~\ref{fig:pure_prediction1}.\footnote{Note that the activation function was assigned as a Python dictionary.}   



So far, we have demonstrated that \rctorchS satisfies two of the three of Bengio's criteria for a successful time-dependent model. The RC works with data for a long duration (many discrete time steps) and trains quickly. However, we have not yet demonstrated its performance on noisy data.




\subsection{Noisy Data}

\begin{figure}[ht!]
  \hspace*{-0.7cm}
  \makebox[\textwidth][c]{\includegraphics[width = 0.7\textwidth]{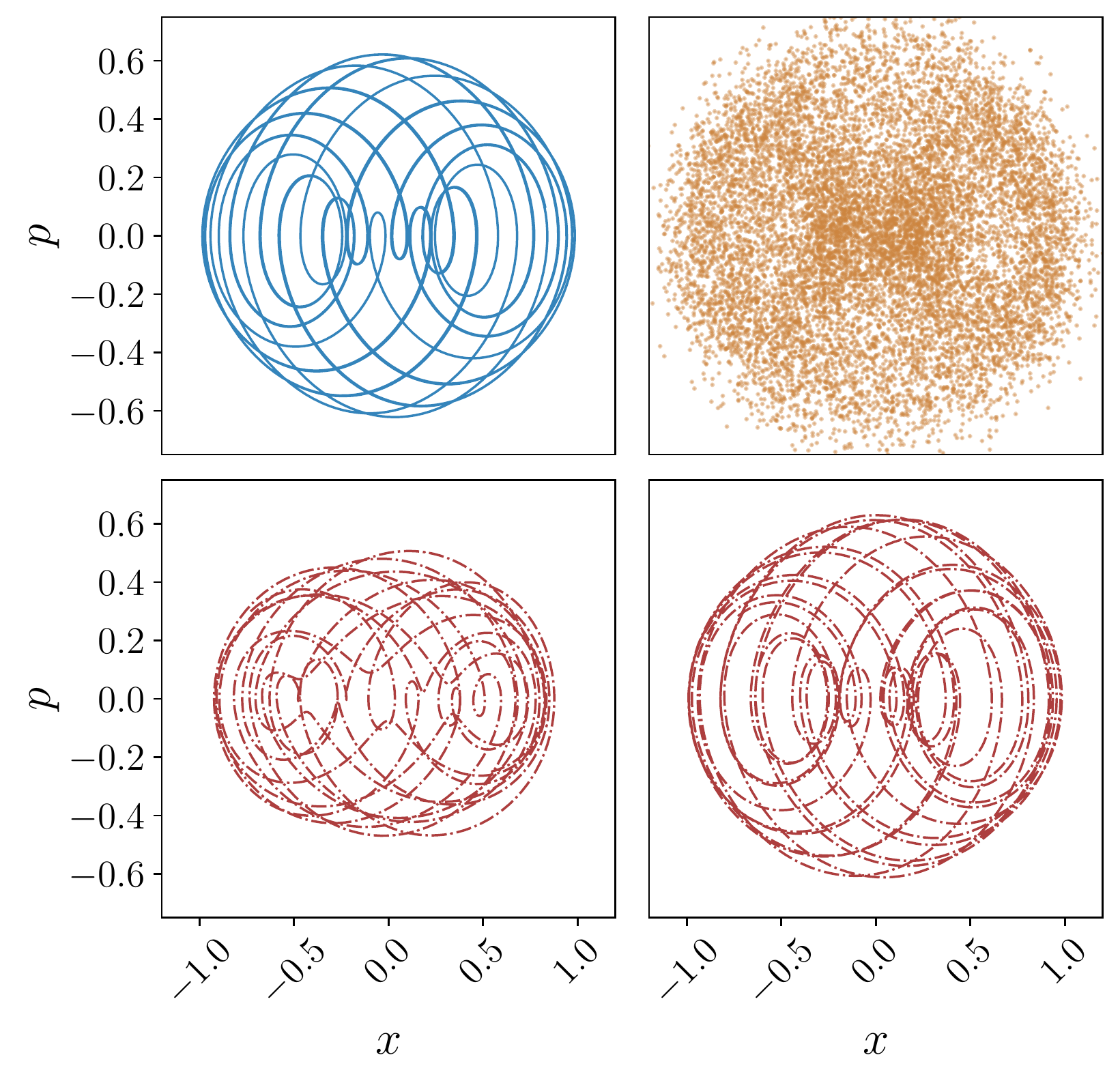}
  }

\caption{Top panel: Ground truth data without (left, blue) and with (right, gold) noise.
Lower panel: 2 RcTorch RcNetworks were trained  on the noisy data. RcTorch RC predictions are plotted in red with a dashed line. The RC prediction plotted at lower left was a pure prediction and the lower right prediction was parameter-aware.}
\label{fig:noise_phase}
\end{figure}

To test our model on noisy data, we selected a new force $f=0.5000  \sin(0.3867 t)$ and generated new data with an \texttt{odeint} integrator. Using a 40-60\% train-test split, we fit an RC on $8000$ time-steps and predict an additional 12000 time-steps with the same HPs as in the previous section. However, this time we added random uniform noise to the training data where $y$ was perturbed by $\epsilon$, which is drawn from a uniform distribution within the range -0.15 and 0.15. In Fig.~\ref{fig:noise_phase} we demonstrate that the RC is able to recover the phase space and therefore was able to discover the underlying dynamics of the system despite a significant amount of noise. This clearly demonstrates the robustness of \rctorchS to noise.\footnote{Note that the parameter-aware RC outperforms the pure-prediction RC as expected.}

\pagebreak

\subsection{Force comparison/Transfer learning}

\label{section:transfer_learning}

Previously, we demonstrated that the RC can solve the driven pendulum for a particular force. In order to demonstrate the robustness of the RC, here we show how the RC performs when the problem is slightly altered.
In general, a set of RC HPs (and the hidden states that result from those HPs) can be seen as a high dimensional embedding of a problem on which the RC is trained. In this section, we show that HPs can be transferred between similar experiments without a significant drop in performance. We generated new data with the \texttt{odeint} integrator for varying values of $\alpha$ and $\omega$ in order to change the driven force of the pendulum.\footnote{ In particular we expressed the force as follows: $f(\alpha, \omega, t) = \alpha * f(\omega  t)$, $f \in \sin(), \cos(), \sin()\cos()$.} Some examples of the effect of changing $\alpha$ and $\omega$ were visualized in the trajectory and phase space plots in Fig. \ref{fig:example_trajs}.

\begin{figure}[ht!]
  \hspace*{-.85cm}
  \makebox[\textwidth][c]{
  \includegraphics[width = 
  0.8\textwidth]
  {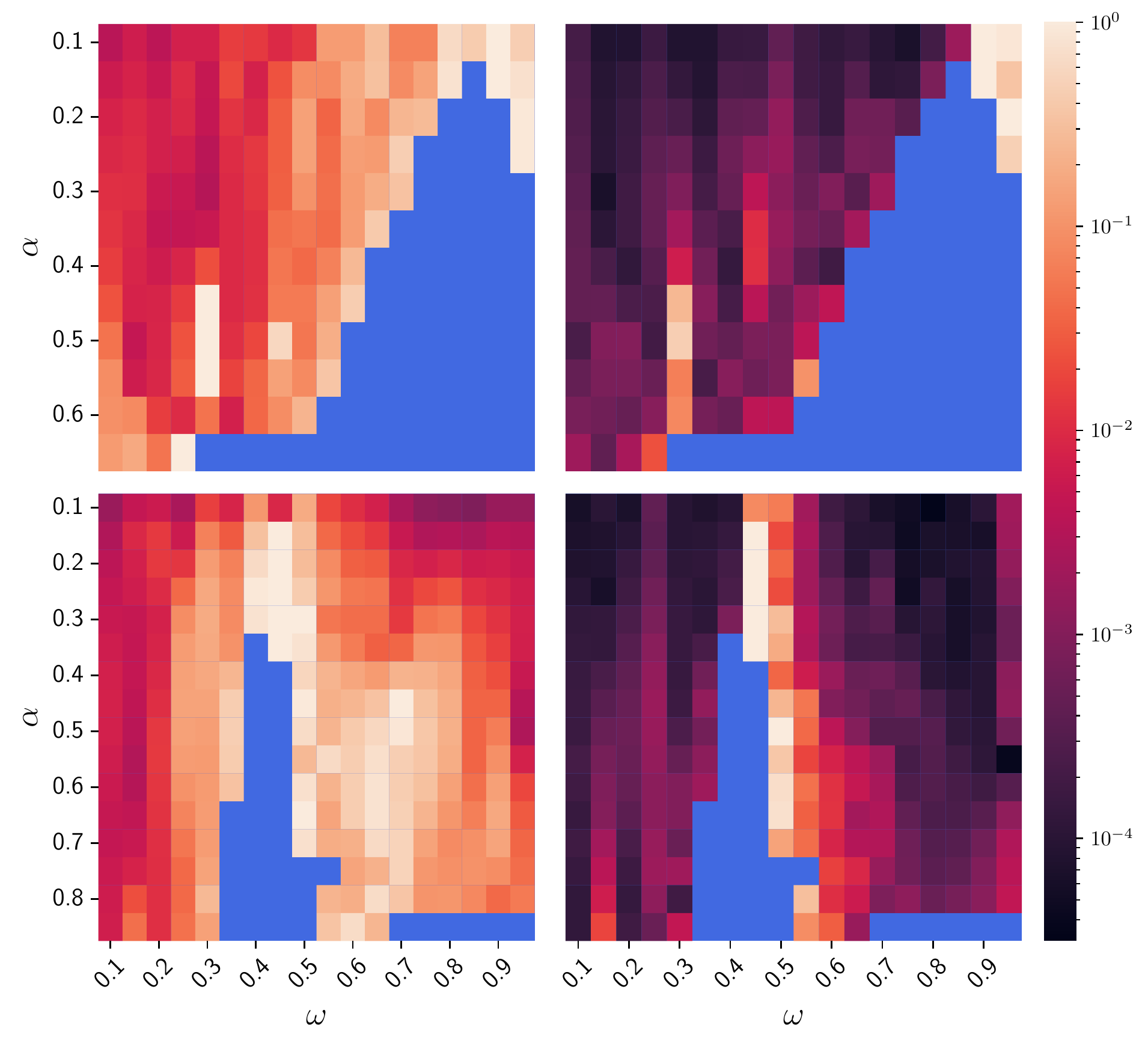}}

\caption{ This figure shows the performance of RCs that are trained on different driven forces of the forced pendulum. The heat-maps in the left hand column (top left and bottom left) plot scores of RCs doing pure prediction. The heat-maps in the right hand column (top right and bottom right) plot the scores of parameter aware RCs. To generate the upper row RC predictions (top left and top right) data was simulated from an integrator using a forced pendulum where the force $f=\alpha \sin\left(\omega t\right)$. The lower row RC predictions (bottom left and bottom right)
were trained based on data generated with force $f=\alpha \sin\left(\omega t\right)\cos\left(\omega t\right)$. The MSE is represented by the red color value that is shown on a logarithmic scale. Resonant frequencies in blue have been removed.}
\label{fig:transfer_learning}
\end{figure}


In Fig.~\ref{fig:transfer_learning} each rectangle represents a trained RC. It is important to note that all these RCs shared the same set of HPs, which were introduced in section \ref{section:hps}. The color of the rectangles corresponds to the MSE of that network, ranging from dark red (low error) to light red (high error). Forces that resulted in resonance (unstable/aperiodic behavior) are colored blue and are ignored. We compare pure prediction RCs (top left and bottom left heat maps) and parameter aware RCs (top right and bottom right heat maps) side by side. It is clear from the darker color values that the parameter-aware RCs perform better. In addition, we manipulate the forces by using two different base equations,
which are $f=\alpha \sin\left(\omega t\right)$ (the top left and top right heat-maps) and $f=\alpha \sin\left(\omega t\right)\cos\left(\omega t\right)$ (the bottom left and bottom right heat-maps) in the first and second rows of Fig.~\ref{fig:transfer_learning}.





Overall, low MSE values represented by large swaths of dark color indicate that the RC is robust to changes in force. This implies that the RC hidden states learn something fundamental about the dynamics of the forced pendulum so that we can conduct transfer learning for similar forces. 
However, there are small regions of the heat-maps with high MSE values. These light colored patches correspond to values of $\alpha$ and $\omega$ for which the RC did not perform as well. These regions, where the loss is larger, are found mostly on the edge of resonant behavior (blue patches), for example $(\alpha = 0.1, \omega = 0.95 ), \text{ force} = sin()$ and $(\alpha = 0.5, \omega = 0.5 ), \text{force} = sin()cos()$. In the next section, we address this issue by solving for new HPs.

\subsection{Bayesian Optimization}
\label{section:BO}

\rctorchS takes some of the best BO methods that have come out of Facebook, Uber, and MIT, and synthesizes them to empower and automate RC. In particular, \rctorchS relies on a custom implementation of the \turboS algorithm in combination with the \texttt{BoTorch} package. \cite{Botorch2020, TURBO2020} We modified the \turboS algorithm outlined in a \texttt{BoTorch} tutorial.\cite{TURBO2020, Botorch2020} \turbo, like global BO methods, is robust to noisy observations and has rigorous uncertainty estimates. However, an advantage of \turboS over other BO algorithms is that it does not suffer from ``the overemphasized exploration that results from global optimization" and does not ``scale poorly to high dimensions".\cite{TURBO2020} Turbo performs well by breaking down the global optimization problem into many smaller local optimization problems. Moreover, Turbo has been shown to be faster than other state-of-the-art black-box evaluation methods and has been shown to find better global maxima.\cite{TURBO2020}

\begin{figure}[ht!]
  \centering
\includegraphics[scale=0.45]{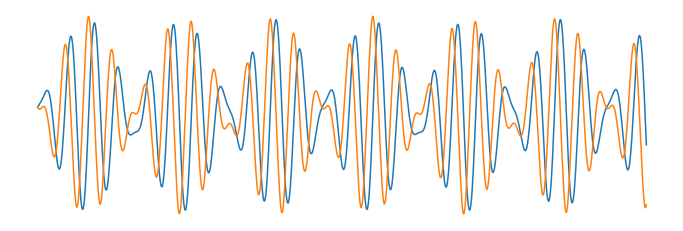}

 \caption{Trajectory plot. The position and velocity are plotted in blue and yellow respectively, of the forced pendulum when $f=0.5\sin(0.6 t)\cos(0.6 t)$. The shape of this trajectory is different enough (based on the transfer learning loss function heat-maps) shown in the previous section, to warrant a new HP search.}
 \label{fig:bot}
\end{figure}

In order to obtain the set of HPs from the previous sections (first introduced in section \ref{section:hps} and used until this point), \rctorchS \rcbayes \space was run on data generated with an integrator where the driven force $f=0.5 \sin(0.2t)$. From Fig. ~\ref{fig:transfer_learning} we notice that the RC performed poorly for $\sin()\cos()$ forces near $\alpha=0.5$ and $\omega=0.6$. These two sets of dynamics are dissimilar enough to warrant the selection of a new set of HPs. An integrator was used to generate new data based on the new equation for the derivative of the momentum of the forced pendulum: $\dot{p} = -sin(x)  + 0.5\sin(0.6 t)\cos(0.6 t)$.\footnote{One problem with RNNs in general is the fact that the timestep interval $dt$ must be fixed. Here, $dt$ was changed from $1/(200\pi)$ to $1/(20\pi)$.} For the initial position $x_0 = 0.1$ and the initial velocity $v_0 = 0.1$ Fig.~\ref{fig:bot} shows the corresponding trajectory.

\begin{figure}[ht!]

  \hspace*{-0.64cm}
  \makebox[\textwidth][c]{\includegraphics[width = 
  0.95\textwidth]{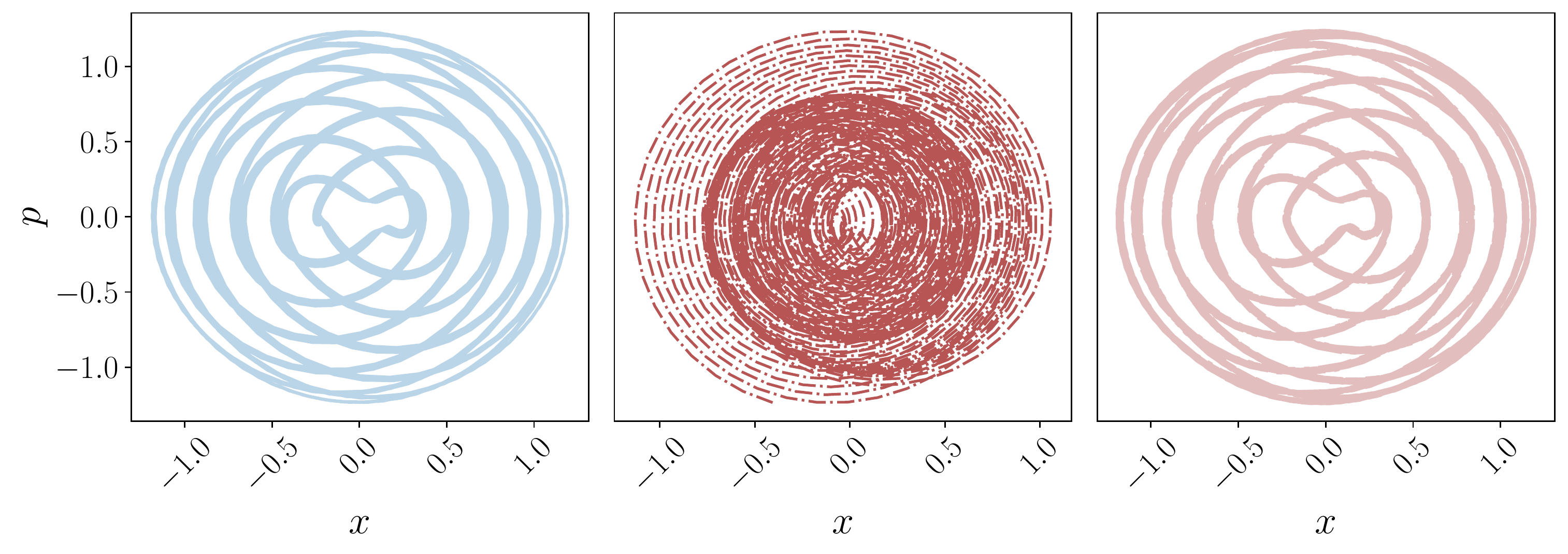}}
 
 \caption{ Phase plots. In the left panel, the ground truth data is plotted for the forced pendulum when $f=0.5\sin(0.6 t)\cos(0.6 t)$. In the center panel, the prediction is plotted from an RC which defined by the HPs from the previous section. In the right panel, the RC prediction with the new re-optimized HPs is plotted. The ground truth data is plotted with a blue line. The RC predictions were plotted in a red dashed line.}
\label{fig:last}
\end{figure}






The \rctorchS \rcbayes \space class does the hard work of tuning the RC's HPs. While at first glance this class may seem very complicated, it is in fact very easy to use. First, the \texttt{bounds\_dict} defines the search bounds within which to search for the HPs. Next, an \rcbayes \space class object is declared. In this example, feedback was used. \rctorchS was instructed to use six localized \turboS arms in parallel by including the \texttt{n\_jobs = 6} argument. The random seeds for the reservoirs to be created were also specified and multiple activation functions were employed.


\begin{center}
\begin{tabular}{c}
\begin{lstlisting}

>>> bounds_dict = {`log_connectivity' : (-2, -0.1), 
               `spectral_radius' : (0.6, 2),
               `n_nodes' : (250, 253),
               `log_regularization' : (-3, 3),
               `leaking_rate' : (0, 1),
               `bias': (0, 1)}
>>> rc_bayes = RcBayesOpt(bounds = bounds_dict, 
                          feedback = True, 
                          scoring_method = `nmse',
                          n_jobs = 6, 
                          initial_samples = 10, 
                          random_seed = 210,
                          activation_function = 
                                {`relu' : 0.33, 
                                 `tanh' : 0.5, 
                                 `sin' : 0.1})

\end{lstlisting}
\end{tabular}
\end{center}

Next, we call the optimize method to run the BO. We run \turboS which will train $n$ parallel local BO arms in parallel.

\begin{center}
\begin{tabular}{c}
\begin{lstlisting}
>>> opt_hps = rc_bayes.optimize( n_trust_regions = 6, 
                              max_evals = 1200,
                              x = input_tr, 
                              y = target_tr)
>>> opt_hps
{`connectivity': 0.34275345999343795,
 `spectral_radius': 1.5731128454208374,
 `n_nodes': 251,
 `regularization': 0.1880535350594487,
 `leaking_rate': 0.03279460221529007,
 `bias': 0.8625065088272095} 
\end{lstlisting}
\end{tabular}
\end{center}

By comparing the phase-space plots in Fig.~\ref{fig:last} the dynamics of the system are clearly well captured. In particular, the 
 $MSE = 0.879$ with the old HPs and with the new HPs the  $MSE = 0.0009$.\footnote{\rctorchS has a useful method called \texttt{recover\_hps} which can recover the best scoring HPs if the BO run hangs. This can be achieved either by directly calling the method if the practitioner is running code in a Jupyter notebook, or by wrapping the BO code with a \texttt{try} \texttt{except} catch.
} 
		






\section{Conclusion}


Reservoir computing holds great promise for the future of AI research and applications. Here, we introduce \rctorchS as the first unified \pytorchS RC library with user-friendly hyper-parameter tuning. The library is written to be updated very easily. 

It is easy to install \rctorchS using the command \texttt{pip install rctorch} and to apply it to a new data set. The associated documentation can be found on our github page: \url{https://github.com/blindedjoy/RcTorch} or at readthedocs \url{https://rctorch.readthedocs.io/en/latest/Pages/tutorials/forced_pendulum.html}. This paper aims to allow researchers all over the world to download \rctorch and use it to rapidly solve problems described by sequential data.


\printbibliography

\pagebreak

\end{document}